\documentclass{isprs}
\usepackage{setspace}
\usepackage{geometry} 
\usepackage{epstopdf}
\usepackage[labelsep=period]{caption}
\usepackage{siunitx}
\usepackage{microtype}

\usepackage{etoolbox}
\robustify\bfseries
\usepackage{upgreek}
\usepackage[hidelinks]{hyperref}
\usepackage{tabularx}
\usepackage{booktabs}
\usepackage{subfig}
\usepackage{multirow, makecell}
\usepackage{color}
\usepackage{natbib}
\usepackage{paralist}
\usepackage{amsmath}
\usepackage[super]{nth}
\usepackage[nameinlink]{cleveref}
\usepackage{float}
\usepackage{stfloats}
\usepackage{soul}
\begin{document}

\title{Estimating Chlorophyll A concentrations of several inland waters with hyperspectral data and machine learning models}

 \author{
 P. M. Maier\textsuperscript{1}, S. Keller\textsuperscript{1}}
 \address{
 	\textsuperscript{1 }Karlsruhe Institute of Technology (KIT), 	Institute of Photogrammetry and Remote Sensing, \\
 	Englerstr. 7, D-76131 Karlsruhe, Germany
	\\(philipp.maier, sina.keller)@kit.edu \\
 }

\commission{I,}{?}
\workinggroup{I/1}
\icwg{}

\abstract{
Water is a key component of life, the natural environment and human health.
For monitoring the conditions of a water body, the chlorophyll~\textit{a} concentration can serve as a proxy for nutrients and oxygen supply.
In situ measurements of water quality parameters are often time-consuming, expensive and limited in areal validity. 
Therefore, we apply remote sensing techniques.
During field campaigns, we collected hyperspectral data with a spectrometer and in situ measured chlorophyll~\textit{a} concentrations of $13$ inland water bodies with different spectral characteristics.
One objective of this study is to estimate chlorophyll~\textit{a} concentrations of these inland waters by applying three machine learning regression models: Random Forest, Support Vector Machine and an Artificial Neural Network.
Additionally, we simulate four different hyperspectral resolutions of the spectrometer data to investigate the effects on the estimation performance.
Furthermore, the application of first order derivatives of the spectra is evaluated in turn to the regression performance.
This study reveals the potential of combining machine learning approaches and remote sensing data for inland waters.
Each machine learning model achieves an $R^2$-score between \SIrange{80}{90}{\percent} for the regression on chlorophyll~\textit{a} concentrations.
The random forest model benefits clearly from the applied derivatives of the spectra.
In further studies, we will focus on the application of machine learning models on spectral satellite data to enhance the area-wide estimation of chlorophyll~\textit{a} concentration for inland waters.
} 
\maketitle

\section{Introduction}
\label{sec:introduction}

\sloppy

Clean and fresh water is a key resource for the environment and human health.
Water quality, however, is threatened by over-fertilization leading to algal blooms, oxygen deficiency and hence, mass death of fish.
Furthermore, some phytoplankton species, especially blue-green algae, can be harmful for humans.
For example, they might spread into drinking water reservoirs and release toxic substances.
To overview and assess the dangerous effects of such algal blooms, a continuous monitoring of the algae growth is advisable.

The feasibility of conventional in-situ monitoring approaches is restricted due to their limitations of both spatial coverage and temporal frequency of the recording.
As a consequence, we approve and follow the remote sensing approach which is already used for monitoring chlorophyll~\textit{a} concentration in water since the 1990’s~\citep{Gitelson.1992}.
Data recorded by remote sensing techniques e.g. satellites is used successfully in the context of estimating chlorophyll~\textit{a} concentration in ocean waters e.g. by ~\cite{OReilly.1998}.
Regarding inland waters, this estimation seems to be more challenging~\citep{Palmer.2015}.
The available multispectral satellite data is primarily useful for observing ocean water or land surface.
For monitoring small water bodies, the spatial resolution of satellite images is often too low.
The same applies for the spectral resolution, which is rather insufficient for e.g. estimating phytoplankton pigments based on satellite data~\citep{Palmer.2015}.
In addition, inland waters are optically more complex than ocean waters because of various suspended particles~\citep{Hunter.2008}. Thus the transferability of the estimation model is not ensured.

The relation between algae existence and remote sensing is primarily linked to the absorption of light in a wavelength of \SI{665}{\nm} on the pigment chlorophyll~\textit{a}~\citep{Morel.1977}.
This reflection minimum and the following reflection maximum around \SI{700}{\nm} serve as the basic wavelengths of the band ratio approaches for the identification of chlorophyll~\textit{a} in inland waters.
Ratio approaches are often used by e.g.~\cite{Gitelson.1992,schalles1998,Gons.1999,DallOlmo.2003,Gitelson.2007,Zhou.2013}.
\cite{schalles1998} rely on the area under the peak around \SI{700}{\nm} or just the respective amplitude for the estimation of chlorophyll~\textit{a} concentration.
Alternatively, derivatives of the spectra are applied in this context as well~\citep{Rundquist.}.

Another approach for the estimation of water parameters by hyperspectral data are data-driven machine learning models primary based on supervised learning.
For supervised learning, a dataset is divided into sets where e.g. one set is used for the training of the model and the other set is employed to evaluate the model.
The potential of these machine learning models for the estimation of several water parameters is shown in the context of coastal waters by e.g.~\cite{Keiner,Gonzales,Kim.2014}, in the context of rivers by~\cite{Maier.2018} and~\cite{Keller.2018} as well as for big lakes by~\cite{Odermatt}.

For the recording of the remote sensing data a multitude of different sensor techniques are applied: spectrometers, hyperspectral cameras and satellite data.
They vary widely with respect to the spatial and spectral resolution.
Satellite data offers many advantages: the temporal repetition rate is constant, they can cover huge areas and in the long run it is a cost-effective solution.
However, the large pixel size might be adverse for small inland waters and the spectral bandwidth is often too coarse.
~\cite{Decker.1992} analyzed the compatibility of the Landsat TM sensor and the SPOT sensor to the chlorophyll~\textit{a} absorption.
Neither of the sensors cover the range between \SI{690}{\nm} and \SI{760}{\nm}, so the peak around \SI{700}{\nm} related to chlorophyll~\textit{a} cannot be detected.
Then again, push broom sensors with narrow bandwidths of \SI{10}{\nm} to \SI{20}{\nm} between \SI{600}{\nm} to \SI{720}{\nm} seem to be practicable for the estimation of phytoplankton substances~\citep{Decker.1992}.
With the launch of the DESIS (DLR Earth Sensing Imaging Spectrometer) mission in 2018 and the upcoming launch of EnMAP (Environmental Mapping and Anlaysis Program), monitoring of inland waters with reasonable spatial extent and hyperspectral resolution should be feasible.
Up to now, unmanned aerial vehicles (UAVs) provide an appropriate spectral and spatial resolution for any remote sensing based monitoring approach over limited areas.

In this study, we assess the transferability of applied machine learning models for the estimation of chlorophyll~\textit{a} concentrations with hyperspectral data and for several inland water bodies.
Therefore, we recorded our own hyperspectral dataset in several field campaigns with a spectrometer\footnote{JB Hyperspectral Devices UG, Germany} from 13 different inland waters.
As reference data, we collected water samples, which were evaluated with a photometer\footnote{AlgaeLabAnalyser bbe moldaenke GmbH, Germany} regarding their chlorophyll~\textit{a} concentration.
Finally, the dataset for this study contains of $422$ datapoints including hyperspectral data and reference data.
For the estimation of chlorophyll~\textit{a} concentration, we apply three different regression models: Random Forest (RF), Support Vector Machine (SVM) and an Artificial Neural Network (ANN).
An important factor for the estimation performance of a model is the spectral resolution of the sensor.
As we rely on a dataset collected with a spectrometer, we aggregate the spectra to several bands with different resolutions to simulate hyperspectral cameras or satellite sensors.
After various pre-tests, we decided to apply four different resolutions with a continuous interval of \SI{4}{\nm}, \SI{8}{\nm}, \SI{12}{\nm} and \SI{20}{\nm}.
Additionally, we calculated derivatives of the different aggregated data and applied the same regression models.

The objectives of this study are:
\begin{compactitem}
    \item to describe the recorded dataset including the measurement setup of our field campaigns,
    \item to demonstrate the potential of supervised learning models for the estimation of chlorophyll~\textit{a} concentration of different inland waters,
    \item to assess the spectral resolution in the context of estimating chlorophyll~\textit{a} concentration as well as to determine, which bandwidths are suitable to achieve a sufficient regression performance,
    \item to measure the effects of using derivatives of a spectrum to estimate chlorophyll~\textit{a} concentrations and
    \item finally to evaluate the different machine learning models.
\end{compactitem}

We describe the applied sensor systems of the field campaigns and the measured dataset in \Cref{sec:dataset}. 
In \Cref{sec:method}, the presentation of the machine learning models follows.
\Cref{sec:results} contains the evaluation of the measured dataset and the assessment of the different approaches.
Finally, we conclude our studies in \Cref{sec:conclusion} and give an overview about future research application based on the presented dataset.

\section{Sensors and Dataset}
\label{sec:dataset}

To reveal the potential of supervised learning models for the estimation of chlorophyll~\textit{a} concentrations in different inland waters, many data of such waters as well as varying chlorophyll~\textit{a} concentrations are needed.
The presented dataset consists of hyperspectral data and chlorophyll~\textit{a} concentrations.
Both types of data are measured with two different sensor systems.
The recordings were challenging, since we needed to produce comparable hyperspectral data with varying daytime over a measurement period of four months.

\subsection{Sensors and Data Acquisition}
\label{sec:data:sub:sensors}

\begin{figure}[h]
	\centering
    \includegraphics[width=0.45\textwidth]{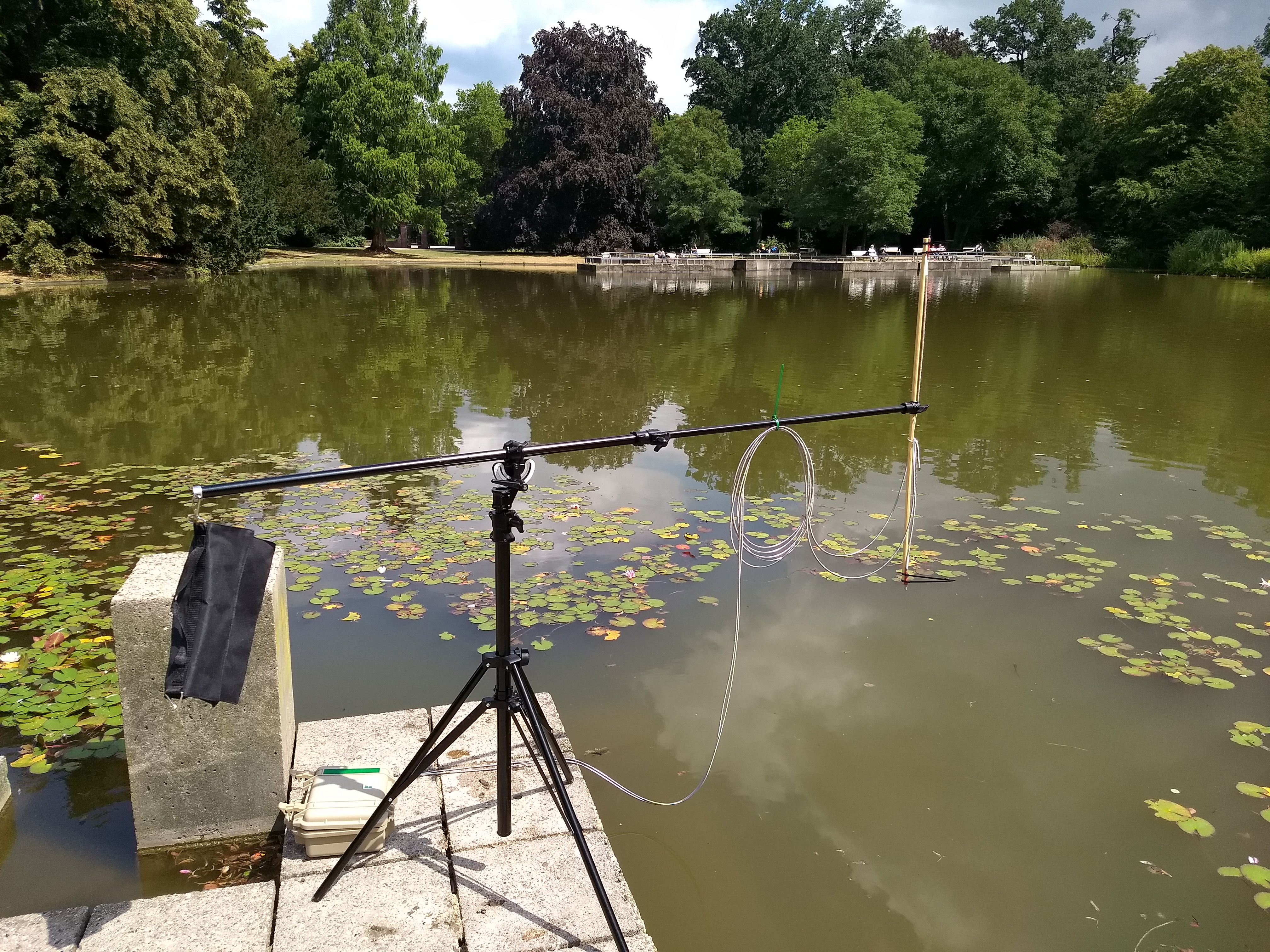}
    \caption{Measurement setup of the RoX spectrometer at an artificial pond.}
    \label{fig:RoXoutside}
\end{figure}

\begin{figure}[h]
	\centering
    \includegraphics[width=0.45\textwidth]{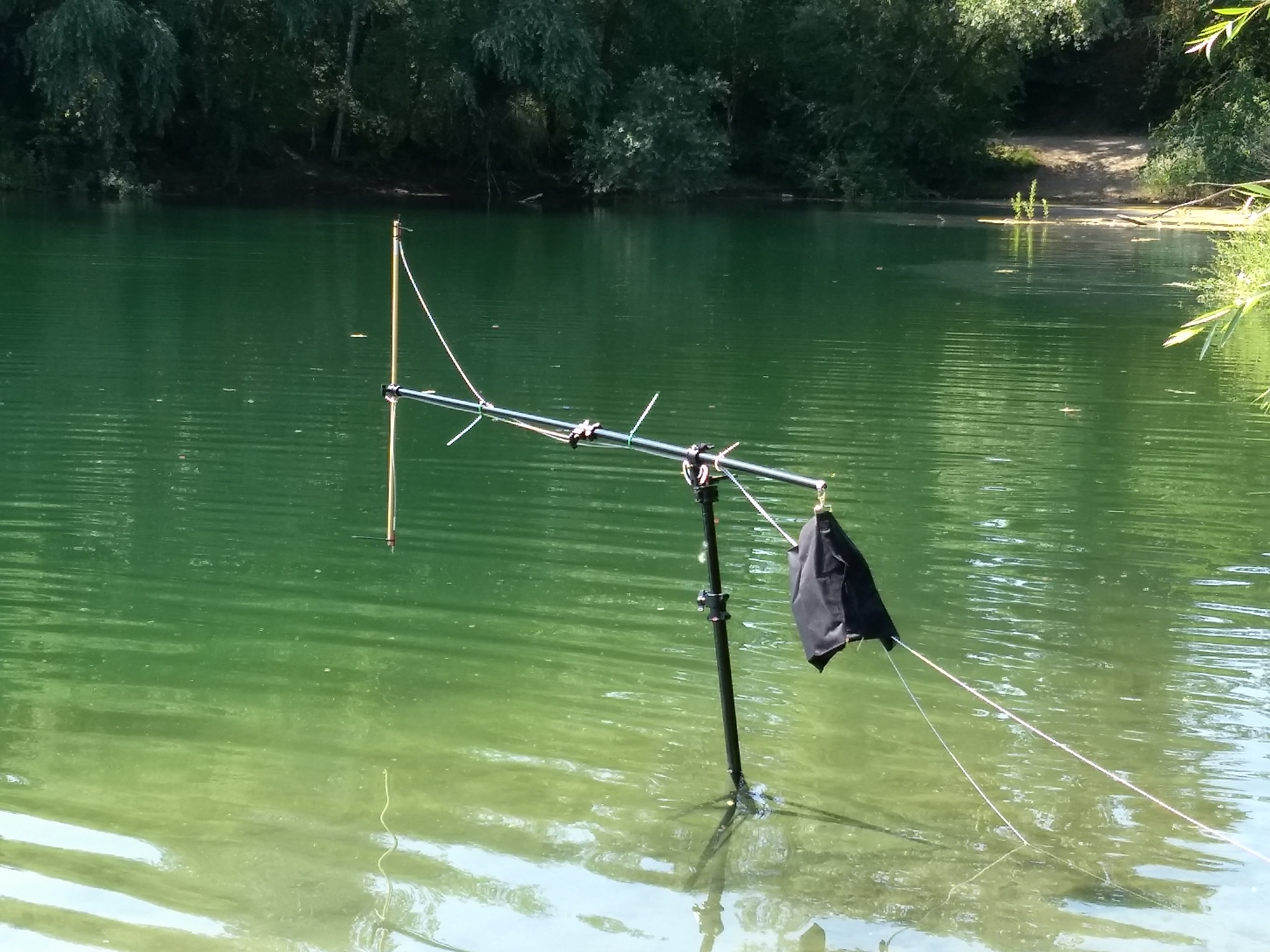}
    \caption{Measurement setup of the RoX spectrometer at a natural water body.}
    \label{fig:RoXinside}
\end{figure}

To record hyperspectral data, we used a so-called Reflectance Box (RoX) spectrometer.
This specific spectrometer covers a spectral range of \SIrange{341}{1015}{\nm} with a sampling interval of about \SI{0.65}{\nm}.
The sensor includes two fiber optic cables, which are oriented in different directions (\Cref{fig:RoXoutside}).
One fiber optic cable is directed upwards and has a cosine receptor at its end. 
This receptor measures the incoming radiation from the sky.
Additionally, it regulates the integration time of the sensor, which is necessary when measuring during different atmospheric conditions such as cloudy conditions or varying sun angle.
The other fiber optic cable is directed downwards to measure the reflectance of the water body and the water surface. 
It has a field of view of \SI{25}{\degree}.
The ratio of these two values yields in a reflectance value in percent, which we use to fit our regression models.
The spectrometer was calibrated in the laboratory with an Ulbricht sphere.

During the field campaigns, the spectrometer was mounted on a tripod to ensure that the cosine receptor was oriented perpendicular to the sky and the other receptor pointed towards the lake surface (see \Cref{fig:RoXoutside}).
The tripod with the spectrometer was placed as far as possible in the water (see \Cref{fig:RoXinside}).
During most of the measurements at natural waters, the lake bed under the spectrometer was invisible.
However, if the water was very clear, it might had been visible.
When measuring at artificial ponds, the tripod was placed outside the water body (see \Cref{fig:RoXoutside}).
The data acquisition took place under clear sky conditions up to the occurrence of cirrus clouds.
The spectrometer's sampling interval during the measurements was adjusted to \SI{15}{\second}.

The water samples for the evaluation with the photometer were collected every five minutes.
They were taken at a depth of \SI{10}{\cm} under the water surface and in an area close to the spectrometer.
The depth of \SI{10}{\cm} was chosen due to two reasons: Firstly, this depth ensured to collect data with the photometer as well as with the spectrometer.
Secondly, regarding the measurements at all water bodies, the \SI{10}{\cm} depth had to be chosen, since in some of the water bodies high chlorophyll~\textit{a} concentrations were often accompanied by high turbidity.
That effect led to a depth of visibility of 20 cm and lower.
Additionally, we took some reference samples in higher depths but the differences in the chlorophyll~\textit{a} concentration compared to the water samples at \SI{10}{\cm} depth were negligible.
Until the water samples were measured in the laboratory, they were protected from sunlight.
The samples were evaluated prompt in a \SI{25}{\ml} cuvette by the photometer, which is able to measure the chlorophyll~\textit{a} concentration in the range from \SI{0}{\micro\gram\per\liter} to \SI{200} {\micro\gram\per\liter}.
These water samples represent the reference data for the chlorophyll~\textit{a} concentrations.

In total, we collected hyperspectral and reference data from $13$ inland water bodies between June and October 2018 in the surrounding area of Karlsruhe, Southwest Germany.
Three of them were artificial ponds and the other ten were natural waters e.g. a branch of a river or flooded gravel pits.
\Cref{tab:overview} provides an overview of the investigated water bodies.
The water bodies were selected by accessibility and proximity to each other to cover different locations within one day.

\begin{table}[tb]
	\centering
	\caption{Description of the investigated water bodies.}
		\begin{tabular}{lSSS}
		\toprule
			\multirow{1}{*}{ID} & {Status} & {Coordinates} & {Description}\\
            \midrule
			A1		& {Artificial}  &   {49.0129N, 8.4104E} & {artificial pond}\\
            A2 	& {Artificial}  &   {49.1312N, 8.4401E} & {artificial pond}\\
            A3		& {Artificial}  &   {49.0170N, 8.4049E} & {artificial lake}\\
            N1     & {Artificial}  &   {49.0385N, 8.3850E} & {flooded gravel pit}\\
            N2     & {Artificial}  &   {49.0999N, 8.3822E} & {branch of a river}\\
            N3     & {Artificial}  &   {49.1074N, 8.3867E} & {flooded gravel pit}\\
            N4     & {Artificial}  &   {49.0396N, 8.4498E} & {flooded gravel pit}\\
            N5     & {Artificial}  &   {49.0514N, 8.4477E} & {small channel}\\
            N6     & {Artificial}  &   {49.0794N, 8.4648E} & {flooded gravel pit}\\
            N7     & {Artificial}  &   {49.0345N, 8.3135E} & {stream}\\
            N8     & {Artificial}  &   {48.9667N, 8.3288E} & {flooded gravel pit}\\
            N9     & {Artificial}  &   {48.9771N, 8.2724E} & {flooded gravel pit}\\
            N10    & {Artificial}  &   {48.9590N, 8.2193E} & {branch of a river}\\
             \bottomrule
		\end{tabular}
	\label{tab:overview}
\end{table}

\subsection{Pre-processing}
\label{sec:data:sub:preprocessing}

We applied several pre-processing steps to prepare the hyperspectral data for the regression on the chlorophyll~\textit{a} concentration.

\begin{compactenum}
    \item The measured hyperspectral data was constrained to the wavelength range of \SIrange{400}{900}{\nm} in order to avoid sensor noise.
    \item Any outlier in the hyperspectral dataset was investigated within the sampling period we measured on a single point.
    Such outlier occur e.g. by sun glint, waves or shadows.
    When datapoints exceeded a certain distance to the median for any wavelengths within the sample interval, we excluded them.
    \item We generated different spectral resolutions by aggregating the spectral bands of the spectrometer to bands with a spectral resolution of \SI{4}{\nm}, \SI{8}{\nm}, \SI{12}{\nm} and \SI{20}{\nm}.
    The obtained spectrum was generated by linear weighting of the spectrometer data.
    \item Additionally, we calculated first order derivatives (from here on: derivatives) of the different aggregated data to generate a further dataset.
    \item Finally, we selected only one of the recorded hyperspectral datapoints, which was within a time span of one minute to the sampled reference data.
    Eventually, one datapoint was defined by the generated hyperspectral bands (see the first pre-processing step) and a chlorophyll~\textit{a} value as reference value.
\end{compactenum}

In total, we obtained $422$ datapoints including hyperspectral and reference data.
Furthermore, we generated two distinct datasets according to the pre-processing step $4$: a raw dataset and a dataset with derivatives, both with different spectral resolutions (see pre-processing step $3$).
The distribution of the chlorophyll~\textit{a} concentration, the reference data,  as well as the distribution of the samples per water body are shown in \Cref{fig:Figure1} and \Cref{fig:Figure2}.

For the supervised machine learning approaches, the pre-processed data was split randomly into two disjunct subsets: the training subset and the test subset. 
The splitting was applied on the distinct datasets (see \cref{sec:data:sub:preprocessing}).
\cite{MaierP.2018} applied a random splitting of $30:70$ on a similar dataset. 
In contrast, we chose a splitting with \SI{50}{\percent} of the data for the training subset and \SI{50}{\percent} of the data for the test subset.
The split ratio is due to a lower number of datapoints in the whole dataset.
However, we needed a sufficient number of datapoints for the training of the models.

\section{Methodology}
\label{sec:method}

\begin{table}[b]
	\centering
	\caption{Hyperparameters of the three regression models which are adjusted by a grid search.}
		\begin{tabular*}{0.46\textwidth}{lcl}
		\toprule
			\multirow{1}{*}{Model} & {{Tuning Parameters}} & {{Meaning}}\\
            \midrule
            \multirow{4}{*}{RF} &  \multirow{2}{*}{mtry} & number of wavelengths   \\
            && considered at each node \\
            & \multirow{2}{*}{min.node.size} & minimum amount of \\
            && datapoints at each node \\
            \midrule
            \multirow{2}{*}{SVM} & {gamma} & {kernel parameter}  \\
            & {cost} & {penalty factor} \\
            \midrule
            \multirow{4}{*}{ANN} & \multirow{2}{*}{size} & number of hidden units \\
            && in the layer \\
            & \multirow{2}{*}{decay} & weighting factor \\
            && of the decay \\
            \bottomrule
		\end{tabular*}
	\label{tab:hyperparameters}
\end{table}

To estimate the chlorophyll~\textit{a} concentration, we selected three different supervised machine learning models: Random Forest (RF)~\citep{Breiman.2001}, Support Vector Machine (SVM)~\citep{vapnik1995the} and an Artificial Neural Network (ANN)~\citep{Ripley.1996}.
We chose those supervised models to investigate the applicability of the machine learning in general.
In addition, the three models have already performed well in estimating water parameters according to~\cite{MaierP.2018} and~\cite{Keller.2018}.

The machine learning models were trained on the training subset by linking the hyperspectral data to the chlorophyll~\textit{a} concentration values.
Hyperparameters and model parameters are characteristic for each regression model. 
The former were chosen before the training phase with a grid search approach while the latter were adapted during the training phase.
For the RF model, extratrees were selected as splitrule due to their best performance in previous studies \citep{MaierP.2018}.
The other hyperparameters of the three different models are summarized and described in \Cref{tab:hyperparameters}.
The SVM was conducted with a radial kernel.

During the training phase, every combination of the grid search was carried out with a $5$-fold cross validation and five repetitions on the training subset for each regression model. 
The combination of the hyperparameters with the best average RMSE performance on the five repetitions was the setup for the final model.

During the test phase, each regression model estimated the chlorophyll~\textit{a} concentration based on the hyperspectral data of the test subset.
The estimated chlorophyll~\textit{a} values were compared to the reference chlorophyll~\textit{a} values.
The coefficient of determination ($R^2$), the root mean squared error (RMSE) and the mean absolute error (MAE) express the estimation performance.

\section{Results and discussion}
\label{sec:results}

\begin{figure}[t]
	\centering
    \includegraphics[width=0.5\textwidth]{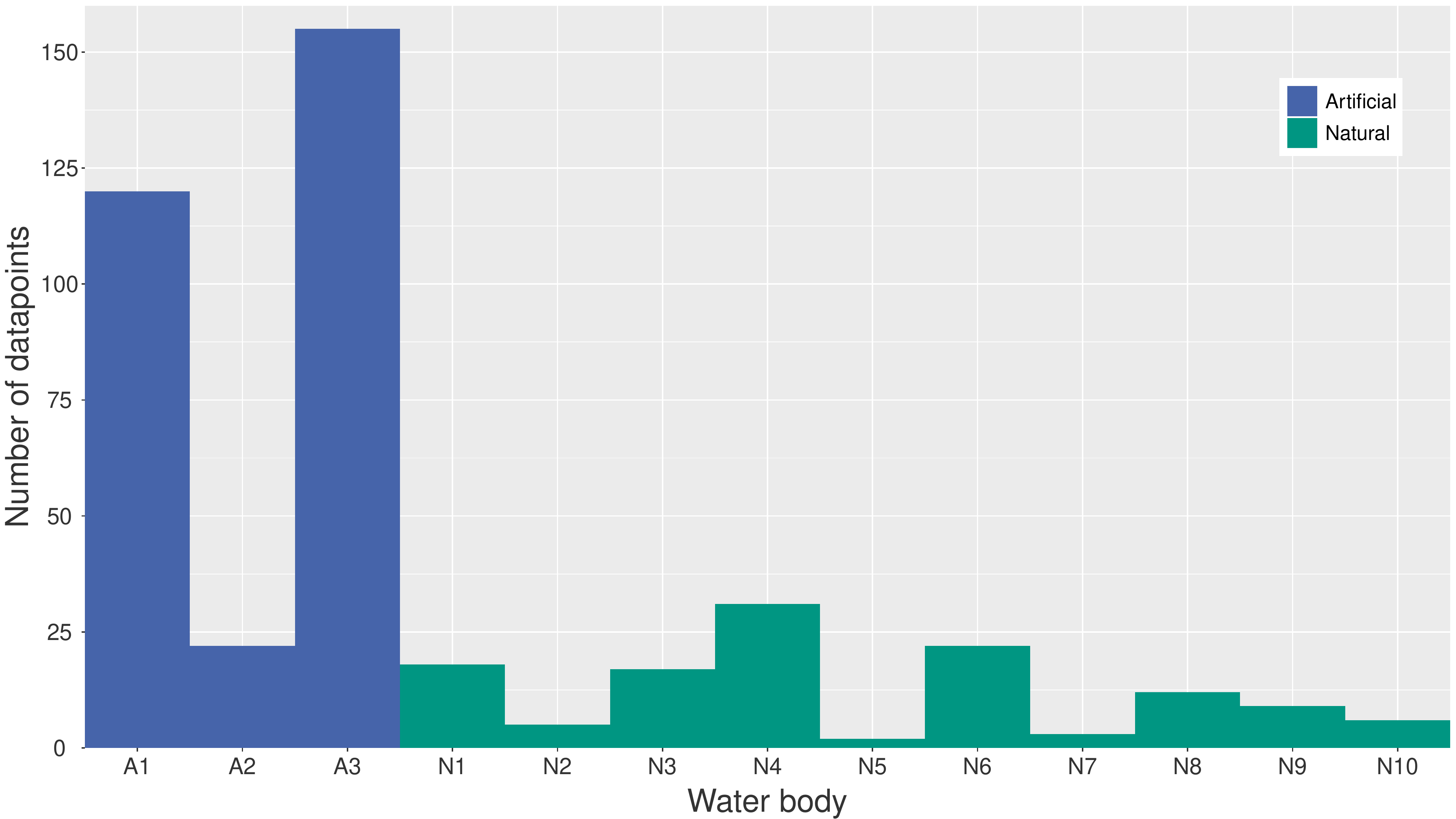}
    \caption{Number of datapoints per water body. The color defines the status of the water body: natural waters (green) and artificial ponds (blue).}\label{fig:Figure2}
\end{figure}

\begin{figure}[t]
	\centering
    \includegraphics[width=0.5\textwidth]{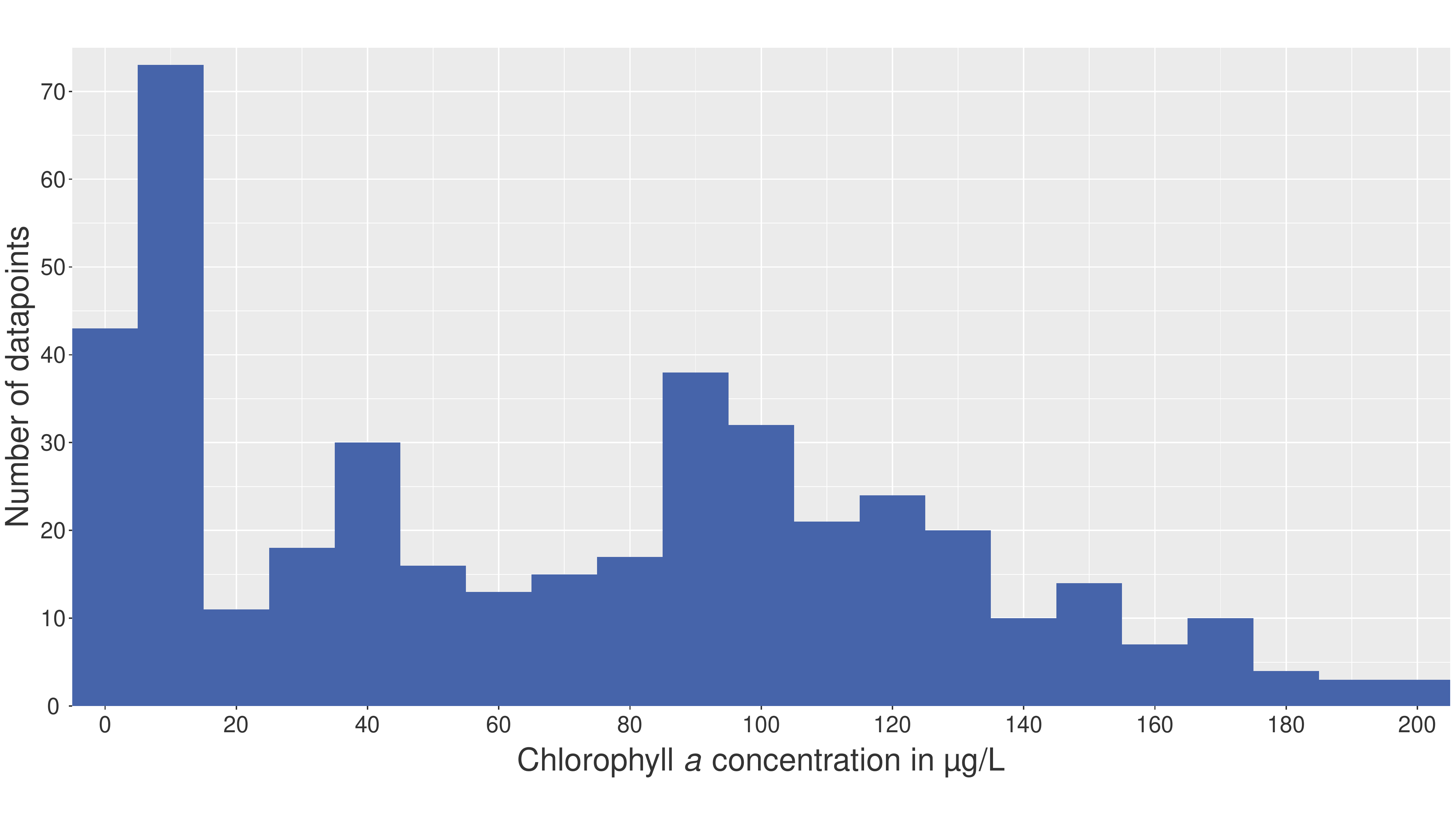}
    \caption{Distribution of the chlorophyll~\textit{a} concentrations in total.}
    \label{fig:Figure1}
\end{figure}

Regarding the distribution of the measured datapoints per location (see \Cref{fig:Figure2}) in detail, it stands out that the number of reference datapoints per location is not uniformly distributed.
Artificial ponds were more frequently examined than natural water bodies.
This is due to the fact, that the natural water bodies in this region are very clean.
There is only a minor variety in the chlorophyll~\textit{a} concentration during the measurement period.
As we noticed, the chlorophyll~\textit{a} concentrations of the artificial ponds, however, varied a lot within the measurement period.
The chlorophyll~\textit{a} concentrations of the investigated water bodies is visualized in \Cref{fig:Figure1}.  
The first three bars show the number of datapoints up to \SI{30}{\micro\gram\per\liter} and represent only the natural water bodies.
While the water samples starting from \SI{30}{\micro\gram\per\liter} belong to artificial ponds.
A uniform distribution of the chlorophyll~\textit{a} concentrations would be optimal for the machine learning models.
This would mean that the same amount of datapoints exist in the training and test subset for every concentration range.
Since we measure under real-world conditions a uniform distribution is not feasible without removing to many datapoints.

\Cref{tab:4nm,tab:8nm,tab:12nm,tab:20nm} present the estimation results of the regression models with different spectral resolution on both datasets, the raw dataset (raw) and the dataset with the derivatives (der).
In general, the best performance is achieved with the highest resolution on both datasets.
However, this does not support the generalization that a lower resolution leads to lower regression performance.
For the best performing regression model on the \SI{20}{\nm}-dataset, the ANN model, the $R^2$ score of \SI{87.1}{\percent} is only \SI{2}{\percent} lower than on the \SI{4}{\nm}-dataset.
The regression results on the \SI{4}{\nm}-dataset and the \SI{8}{\nm}-dataset are rather similar.
Comparing the regression performance on the \SI{20}{\nm}-dataset and on the \SI{12}{\nm}-dataset, the regression  performance on the former exceeds the latter.
A possible reason for this finding could be a better positioning of the bands in terms of chlorophyll~\textit{a} sensitivity for the \SI{20}{\nm}-dataset compared to the \SI{12}{\nm}-dataset.

Regarding the effects of derivatives by comparing the upper and the lower half of the \Cref{tab:4nm,tab:8nm,tab:12nm,tab:20nm}, we can observe that the RF model experiences the strongest influence by derivatives.
Improvements between \SI{5}{\percent} and \SI{10}{\percent} in the $R^2$ score for all resolutions are reached.
For the SVM, we notice slight improvements three times by different resolution with the derivatives, but the effect is not clear.
Applying the ANN model on the dataset with the derivatives, the effect is reversed compared to the SVM due to the specific characteristics of an ANN.
In summary, calculating derivatives of the spectra results in a loss of the absolute reflectance value.
It seems that this circumstance does not influence strongly the estimation performance of the models in general.

Comparing the estimation results between the three machine learning models, ANN shows the best performance for the raw bands and overall.
The best performance is achieved for the \SI{8}{\nm}-dataset with $R^2$ of \SI{89.2}{\percent}.
RF and SVM demonstrate a slightly worse estimation performance.
Both models are in a range of \SI{1}{\percent} $R^2$ to each other for every spectral resolution.

\begin{table}[t]
    \centering    
    \caption{Regression results for chlorophyll~\textit{a} estimation with \SI{4}{\nm} spectral resolution for the raw dataset (raw) and the dataset with derivatives (der).}
    \begin{tabular}{llSSS}
        \toprule
        &\multirow{2}{*}{Model}& {$R^2$} & {RMSE} & {MAE}\\
        && {in \si{\percent}} & {in \si{\micro\gram\per\liter}} & {in \si{\micro\gram\per\liter}}\\
        \midrule
        \multirow{3}{*}{raw}    & RF & 81.7 & 22.5 & 15.9 \\
             & SVM & 86.8 & 19.1 & 13.5 \\
            & ANN & 89.0 & 17.7 & 12.0 \\
        \midrule
        \multirow{3}{*}{der}    & RF & 87.1 & 19.0 & 12.9 \\
             & SVM & 85.2 & 20.5 & 14.9 \\
            & ANN & 85.2 & 20.7 & 14.8 \\
        \bottomrule
    \end{tabular}
    \label{tab:4nm}
\end{table}

\begin{table}[t]
    \centering    
    \caption{Regression results for chlorophyll~\textit{a} estimation with \SI{8}{\nm} spectral resolution for the raw dataset (raw) and the dataset with derivatives (der).}
    \begin{tabular}{llSSS}
        \toprule
        &\multirow{2}{*}{Model}& {$R^2$} & {RMSE} & {MAE}\\
        && {in \si{\percent}} & {in \si{\micro\gram\per\liter}} & {in \si{\micro\gram\per\liter}}\\
        \midrule
        \multirow{3}{*}{raw}    & RF & 77.5 & 24.3 & 16.5 \\
             & SVM & 88.1 & 17.8 & 12.3 \\
            & ANN & 88.3 & 17.5 & 12.1 \\
        \midrule
        \multirow{3}{*}{der}    & RF & 87.8 & 18.1 & 12.0 \\
             & SVM & 88.6 & 17.5 & 12.5 \\
            & ANN & 89.2 & 17.0 & 11.9 \\
        \bottomrule
    \end{tabular}
    \label{tab:8nm}
\end{table}

\begin{table}[t]
    \centering    
    \caption{Regression results for chlorophyll~\textit{a} estimation with \SI{12}{\nm} spectral resolution for the raw dataset (raw) and the dataset with derivatives (der).}
    \begin{tabular}{llSSS}
        \toprule
        &\multirow{2}{*}{Model}& {$R^2$} & {RMSE} & {MAE}\\
        && {in \si{\percent}} & {in \si{\micro\gram\per\liter}} & {in \si{\micro\gram\per\liter}}\\
        \midrule
        \multirow{3}{*}{raw}    & RF & 76.4 & 25.5 & 17.2 \\
             & SVM & 82.4 & 22.2 & 14.5 \\
            & ANN & 82.6 & 22.1 & 14.6 \\
        \midrule
        \multirow{3}{*}{der}    & RF & 83.3 & 21.4 & 13.3 \\
             & SVM & 83.5 & 22.0 & 14.6 \\
            & ANN & 81.5 & 22.9 & 13.9 \\
        \bottomrule
    \end{tabular}
    \label{tab:12nm}
\end{table}

\begin{table}[t]
    \centering    
    \caption{Regression results for chlorophyll~\textit{a} estimation with \SI{20}{\nm} spectral resolution for the raw dataset (raw) and the dataset with derivatives (der).}
    \begin{tabular}{llSSS}
        \toprule
        &\multirow{2}{*}{Model}& {$R^2$} & {RMSE} & {MAE}\\
        && {in \si{\percent}} & {in \si{\micro\gram\per\liter}} & {in \si{\micro\gram\per\liter}}\\
        \midrule
        \multirow{3}{*}{raw}    & RF & 79.4 & 24.4 & 16.9 \\
             & SVM & 81.4 & 23.2 & 15.2 \\
            & ANN & 87.1 & 19.4 & 13.2 \\
        \midrule
        \multirow{3}{*}{der}    & RF & 84.7 & 20.9 & 12.5 \\
             & SVM & 85.2 & 20.7 & 14.3 \\
            & ANN & 87.1 & 19.5 & 13.0 \\
        \bottomrule
    \end{tabular}
    \label{tab:20nm}
\end{table}

\begin{figure}[H]
	\centering
    \includegraphics[width=0.45\textwidth]{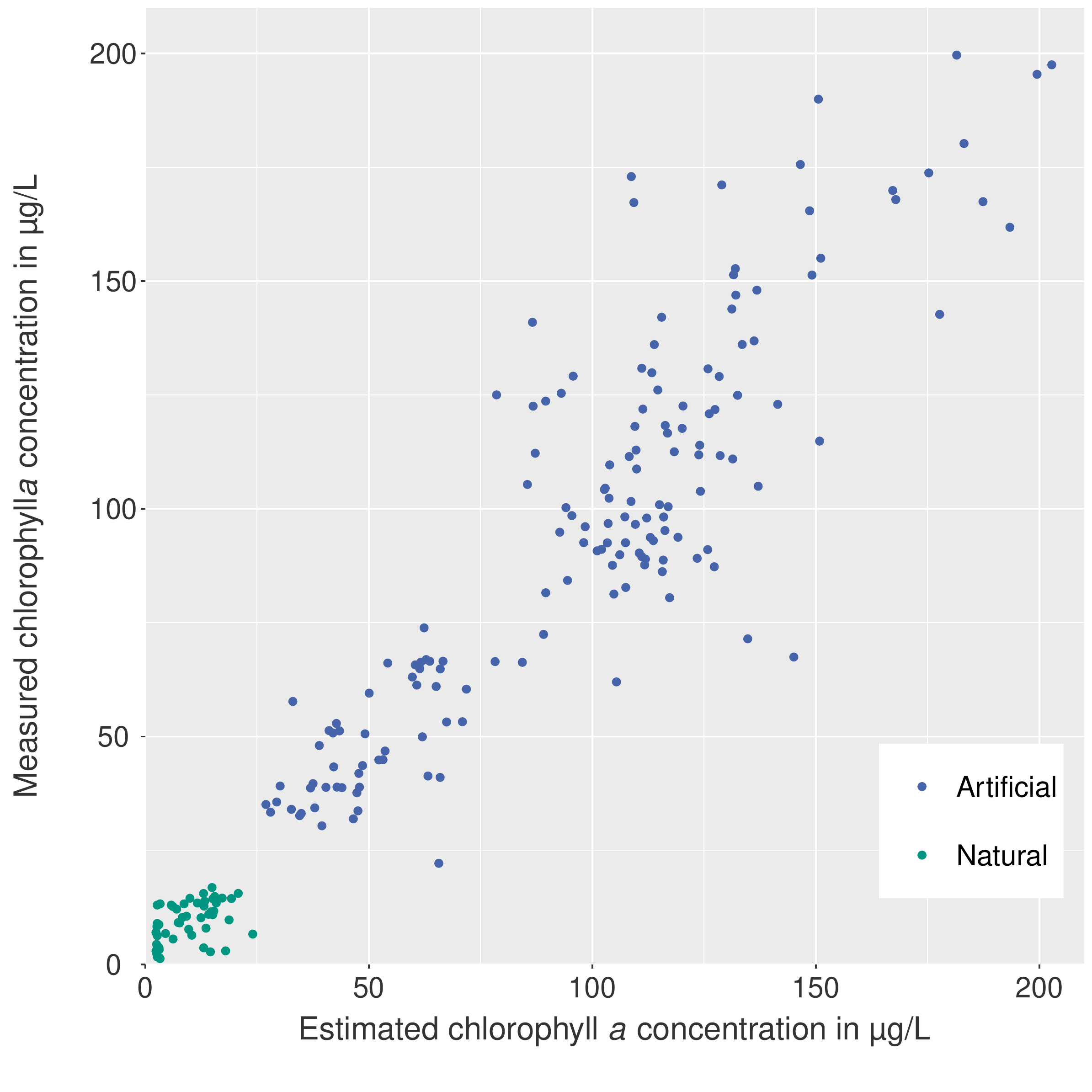}
    \caption{Examplary scatterplot of the ANN regressor showing the estimated vs. the measured chlorophyll~\textit{a} concentration for the \SI{4}{\nm} spectral resolution. The color defines the status of the water body: natural waters (green) and artificial ponds (blue).}
    \label{fig:scatterplot}
\end{figure}

\Cref{fig:scatterplot} visualizes the estimation result of the ANN model for every datapoint of the test dataset with the \SI{4}{\nm} spectral resolution.
In general, we can recognize the regression line distinctly.
However, some points exist, which are estimated poorly.
The green points with high chlorophyll~\textit{a} concentrations represented by the third bar in \Cref{fig:Figure1} are estimated lower than they are measured.
A reason for this aspect could be the small amount of datapoints in this specific concentration range combined with the small amount of samples from the respective water body.
The chlorophyll~\textit{a} concentration range around \SI{200}{\micro\gram\per\liter} contains few datapoints as well.

In summary, the machine learning models show satisfying results in estimating chlorophyll~\textit{a} concentrations by hyperspectral input data for different water bodies.
In contrast to Keller et al. (2018), we obtain a slightly worse estimation performance.
Although, the underlying task of this study has been more challenging since the data was measured in several different water bodies with widely varying visible depth and chlorophyll~\textit{a} concentrations.

\section{Conclusion and Outlook}
\label{sec:conclusion}

In this study, we present a dataset consisting of hyperspectral data and chlorophyll~\textit{a} values, which we measured at $13$ different inland waters.
One main objective is to link the hyperspectral data and the chlorophyll~\textit{a} target values with machine learning models.
The estimation performance of the three applied supervised models RF, SVM and ANN show satisfying results for all the datasets.
With respect to the different spectral resolutions of \SI{4}{\nm}, \SI{8}{\nm}, \SI{12}{\nm} and \SI{20}{\nm}, which we created from the spectrometer data, the machine learning models have even the ability to estimate the chlorophyll~\textit{a} concentration on the dataset with the lowest spectral resolution of \SI{20}{\nm}.
With the perspective of the upcoming EnMAP-mission which has a similar spectral resolution, the opportunity to monitor inland water bodies based on a combination of hyperspectral data and machine learning techniques is demonstrated.
The regression results of the RF model are improved noticeably, by using a dataset with derivatives as input data.
In general, the ANN models show the best estimation performance on the chlorophyll~\textit{a} concentration.

The study confirms the promising results from previous studies \citep{Odermatt,MaierP.2018,Keller.2018}, combining machine learning models with hyperspectral data to estimate chlorophyll~\textit{a} concentrations of inland waters.
Additional, we have faced the challenge to estimate chlorophyll~\textit{a} concentrations of several inland water bodies with varying spectral characteristics satisfactorily.

In future studies, we are planning to conduct further field campaigns to expand the presented dataset.
Additionally, we will focus on the estimation of further water contents such as colored dissolved organic matter or cyanobacteria with hyperspectral data.
Another challenging task could be the up-scaling of the presented methodologies by using data provided by the DESIS Sensor or Sentinel-2 multispectral data.

\section*{Acknowledgement}
\label{sec:Acknowledgement}

The research is part of the WAQUAVID project funded by the German Federal Ministry of Education and Research. We thank our project partners Christian Moldaenke and Andre Zaake for providing the AlgaeLabAnalyser during the whole field campaign, Philipp Wagner and Eva-Maria Klier for their help regarding the measurements as well as Stefan Hinz for his support.

{\footnotesize
	\begin{spacing}{0.9}
		\bibliography{Literatur}
	\end{spacing}
}

\bibliographystyle{isprs}

\end{document}